\documentclass[runningheads, hidelinks]{llncs}
\usepackage{graphicx}

\usepackage{orcidlink}

\usepackage{amsmath,graphicx}

\usepackage{subcaption} %

\usepackage{cleveref} %

\usepackage{cite}
\usepackage{amsmath,amssymb,amsfonts}
\usepackage{mathtools}
\usepackage{commath}
\usepackage{algorithmic}
\usepackage{graphicx}
\usepackage{textcomp}
\usepackage{xcolor}

\usepackage{booktabs}
\usepackage{array}
\definecolor{mygreen}{rgb}{0, 0.56, 0}

\usepackage{wrapfig}

\usepackage[misc,geometry]{ifsym}

\usepackage{float}
\floatstyle{plaintop}
\restylefloat{table}

\newcommand\extrafootertext[1]{%
    \bgroup
    \renewcommand\thefootnote{\fnsymbol{footnote}}%
    \renewcommand\thempfootnote{\fnsymbol{mpfootnote}}%
    \footnotetext[0]{#1}%
    \egroup
}

\begin{document}
\title{JointViT: Modeling Oxygen Saturation Levels with Joint Supervision on 
Long-Tailed OCTA}

\author{Zeyu Zhang\inst{1,2,*}\orcidlink{0009-0006-8819-3741} \and
Xuyin Qi\inst{3}\orcidlink{0009-0000-3172-5475} \and
Mingxi Chen\inst{4}\orcidlink{0009-0002-0286-4630} \and
Guangxi Li\inst{5}\orcidlink{0009-0004-1677-2213} \and
Ryan Pham\inst{3}\orcidlink{0000-0002-4915-097X} \and
Ayub Qassim\inst{1}\orcidlink{0000-0002-2911-2941} \and
Ella Berry\inst{1}\orcidlink{0000-0001-6128-683X} \and
Zhibin Liao\inst{3}\orcidlink{0000-0001-9965-4511} \and
Owen Siggs\inst{1}\orcidlink{0000-0003-2840-4851} \and
Robert Mclaughlin\inst{3}\orcidlink{0000-0001-6947-5061} \and
Jamie Craig\inst{1}\orcidlink{0000-0001-9955-9696} \and
Minh-Son To\inst{1}\orcidlink{0000-0002-8060-6218}}

\authorrunning{Zhang et al.}

\titlerunning{JointViT}

\institute{Flinders University \and
The Australian National University \and
The University of Adelaide \and
Guangdong Technion - Israel Institute of Technology \and
The University of Sydney}

\maketitle              %

\begin{center}
\vspace{-0.6cm}
  \href{https://steve-zeyu-zhang.github.io/JointViT}{\textcolor{blue}{\textbf{\fontsize{7}{4}\selectfont https://steve-zeyu-zhang.github.io/JointViT}}}
\end{center}

\begin{figure}[H]
    \centering
    \includegraphics[width=\linewidth]{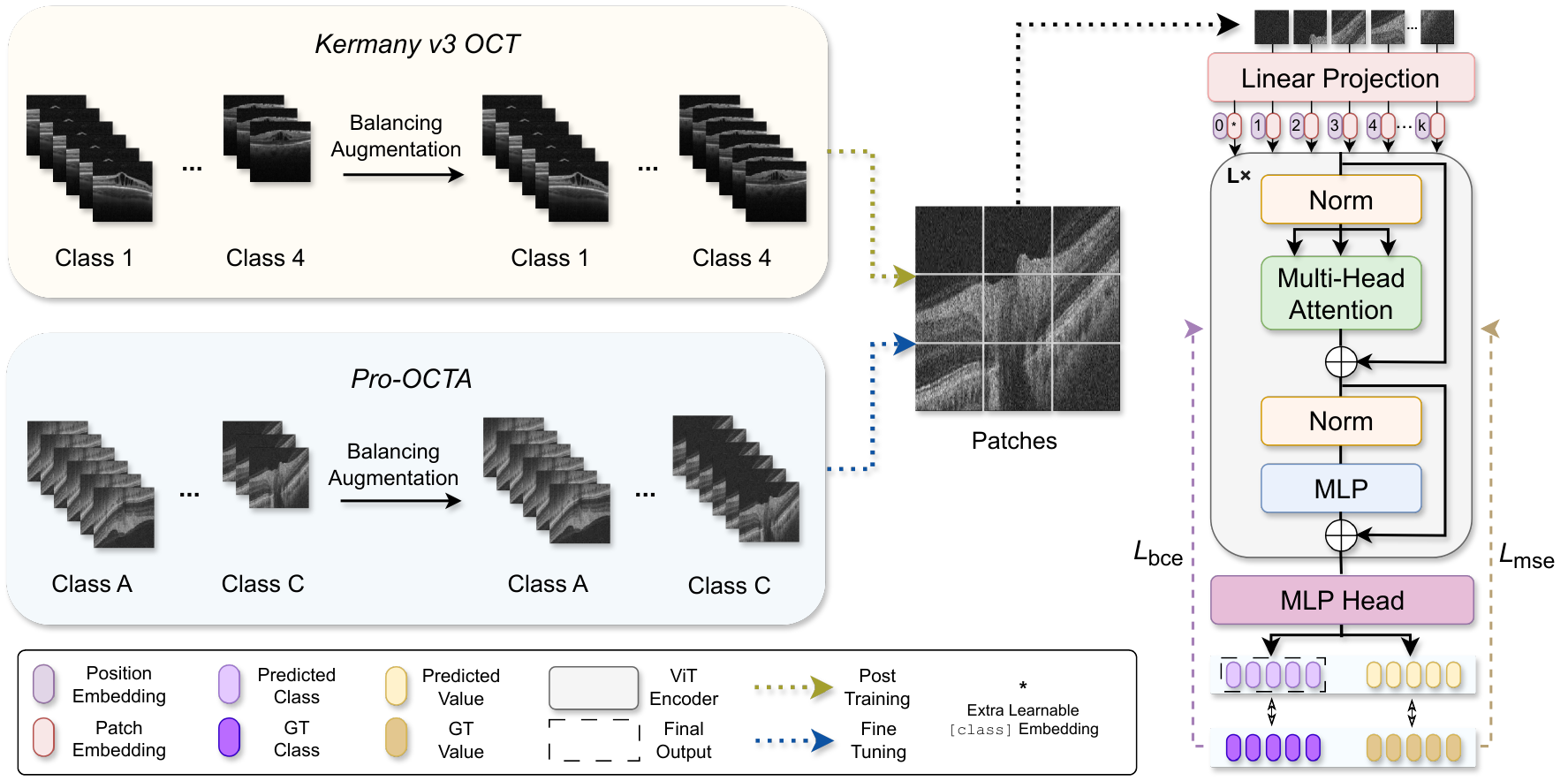} 
    \caption{The figure illustrates the pipeline of our proposed \textbf{JointViT}, which comprises a \textbf{balancing augmentation} and a plain Vision Transformer with a \textbf{joint loss} for supervision. The classes are denoted as numbers in Kermany v3 dataset \cite{kermany2018identifying} and alphabets in Prog-OCTA. \textit{GT} abbreviates ground truth.}
    \label{fig:jointvit}
\end{figure}

\extrafootertext{\scriptsize $^{*}$Work done while being a student researcher at Flinders Health and Medical Research Institute, Flinders University.}

\vspace{-1cm}
\begin{abstract}

The oxygen saturation level in the blood ($\text{SaO}_\text{2}$) is crucial for health, particularly in relation to sleep-related breathing disorders. However, continuous monitoring of $\text{SaO}_\text{2}$ is time-consuming and highly variable depending on patients' conditions. Recently, optical coherence tomography angiography (OCTA) has shown promising development in rapidly and effectively screening eye-related lesions, offering the potential for diagnosing sleep-related disorders. To bridge this gap, our paper presents three key contributions. Firstly, we propose \textbf{JointViT}, a novel model based on the Vision Transformer architecture, incorporating a \textbf{joint loss} function 
for supervision. Secondly, we introduce a \textbf{balancing augmentation} technique during data preprocessing to improve the model's performance, particularly on the long-tail distribution within the OCTA dataset. Lastly, through comprehensive experiments on the OCTA dataset, our proposed method significantly outperforms other state-of-the-art methods, achieving improvements of up to \textbf{12.28\%} in overall accuracy. This advancement lays the groundwork for the future utilization of OCTA in diagnosing sleep-related disorders.

\end{abstract}

\section{Introduction}

Microvascular dysfunction describes a constellation of vessel and endothelial changes including vessel destruction, abnormal vasoreactivity, and thrombosis that lead to tissue damage and progressive organ failure. Microvascular disease has been linked to an increased risk of stroke, cognitive decline, retinopathy, vascular nephropathy and heart failure. Since microvascular dysfunction seen in specific organs may be a manifestation of systemic disease, studying the microvasculature of the retina provides a window into systemic disease, and may serve as a surrogate marker of cumulative injury to the systemic microvasculature [1, 2]. Early detection of microvascular disease provides an opportunity for timely intervention and prevention of future complications. 

Obstructive sleep apnoea (OSA) is the most common sleep-related disorder [3]. Microvascular endothelial dysfunction in OSA driven by vascular stresses during sleep has been linked to the development of hypertension and atherosclerosis, and treatment of OSA with nocturnal continuous positive airway pressure has been shown to improve endothelial function, improve patient’s sleep-related symptoms, and reduce the risk of future cardiovascular morbidity [4]. Untreated, OSA is associated with adverse outcomes such as cardiovascular and cerebrovascular disease, cognitive decline, and depression. 

Low blood oxygen levels during sleep may indicate the presence of OSA or a sleep-related breathing disorder \cite{orabona2024sleep}. 
Oxygen saturation level in the blood ($\text{SaO}_\text{2}$) indicates the percentage of hemoglobin binding sites in the bloodstream occupied by oxygen molecules \cite{hafen2022oxygen}. This metric serves as a crucial indicator of the body's ability to adequately oxygenate tissues and organs \cite{rhodes2022physiology}. 
Typically, $\text{SaO}_\text{2}$ is acquired by a pulse oximeter, which can monitor blood oxygen levels continuously during sleep \cite{li2023detecting}. 
However, as a non-invasive method, it may yield results with some variations and errors that may not fully reflect actual oxygen levels \cite{okunlola2022pulse}. 
Conducting sleep studies also involves monitoring patients all night long with the help of these devices \cite{okunlola2022pulse}. This process is time-consuming and demands continuous monitoring in a clinical setting. 

Optical coherence tomography (OCT) is a non-invasive imaging test that utilises infrared light to obtain 2D and 3D visualisation of structures with micrometre resolution. 
OCT angiography (OCTA) is an extension of the OCT technique that extracts blood flow information of the retina, which allows both vessel structure and haemodynamic parameters to be measured [5] \cite{spaide2018optical}. 
OCTA scans can be acquired in a matter of seconds non-invasively without the use of an intravenous contrast \cite{de2015review}.
Unlike OCT, which primarily provides structural images of the eye, OCTA allows for the visualization of blood flow within the retina and choroid without the need for contrast agents \cite{ferrara2016investigating}. 

The ability of OCTA to image the microvasculature of the retina opens the door to screening not only for sleep-related disordered conditions but also for various other systemic conditions \cite{venkatesh2021association}. However, since the OCTA data is acquired in hospitals where there are more patients with sleep-related disorders and fewer healthy instances, this leads to a long-tailed distribution, making the prediction considerably challenging.
Moreover, directly modeling concrete $\text{SaO}_\text{2}$ values based on a limited and long-tailed OCTA dataset appears to be impractical. 
Instead, predicting $\text{SaO}_\text{2}$ categories is much more robust and will benefit the diagnosis of sleep-related breathing disorders.
In pursuit of this objective, we have proposed a novel approach grounded in Vision Transformer \cite{dosovitskiy2020image} architecture to leverage OCTA data for predicting $\text{SaO}_\text{2}$ categories. Our contributions can be succinctly summarized in the following three points.

\begin{itemize}
    \item We proposed a novel model named \textbf{JointViT} built upon the plain Vision Transformer architecture, which incorporates a well-designed \textbf{joint loss} function that leverages both $\text{SaO}_\text{2}$ categories and $\text{SaO}_\text{2}$ values for supervision.
    \item We introduced a \textbf{balancing augmentation} technique within the data preprocessing phase to enhance the model's performance specifically on the long-tail distribution within the OCTA dataset.
    \item We further conducted comprehensive experiments on the OCTA dataset, which demonstrated that our proposed method has significantly outperformed other state-of-the-art methods, yielding improvements of up to \textbf{12.28\%} in accuracy. This advancement lays the groundwork for the future utilization of OCTA in diagnosing sleep-related disorders.
\end{itemize}

\section{Related Works}

\paragraph{\textbf{Medical Imaging Recognition}} Recognizing medical imaging is a crucial and fundamental task within medical image analysis, playing a vital role in computer vision for medical imaging. The long-term goal of the medical imaging community is to design an effective model architecture and develop an efficient learning strategy to acquire a robust representation of medical imaging. M3T \cite{jang2022m3t} proposed a synergistic method combining 3D CNN, 2D CNN, and Transformer \cite{vaswani2017attention} to accurately classify Alzheimer's disease in 3D MRI images, achieving the highest performance in multi-institutional validation databases and demonstrating the feasibility of efficiently integrating CNN and Transformer for 3D medical images. MedViT \cite{manzari2023medvit} proposed a hybrid CNN-Transformer method to enhance robustness against adversarial attacks and improve generalization ability in medical image diagnosis. FCNlinksCNN \cite{qiu2020development} proposed a deep learning method to delineate unique Alzheimer's disease signatures from multimodal inputs, integrating MRI, age, gender, and Mini-Mental State Examination score. It achieves accurate diagnosis by linking a fully convolutional network with a multilayer perceptron, producing precise visualizations of individual Alzheimer's disease risk. MedicalNet \cite{chen2019med3d} proposed a heterogeneous 3D network method based on 3D-ResNet \cite{he2016deep} to extract general medical three-dimensional (3D) features, achieving accelerated training convergence and improved accuracy in various 3D medical imaging tasks, including lung segmentation, pulmonary nodule classification, and liver segmentation. COVID-ViT \cite{gao2022covid} proposed 2D and 3D forms of Vision Transformer \cite{dosovitskiy2020image} deep learning architecture to classify COVID from non-COVID based on 3D CT lung images, achieving superior performance compared to DenseNet \cite{huang2017densely}.

\begin{figure}[H]
    \centering
    \begin{subfigure}[b]{0.25\textwidth}
        \centering
        \includegraphics[width=\linewidth]{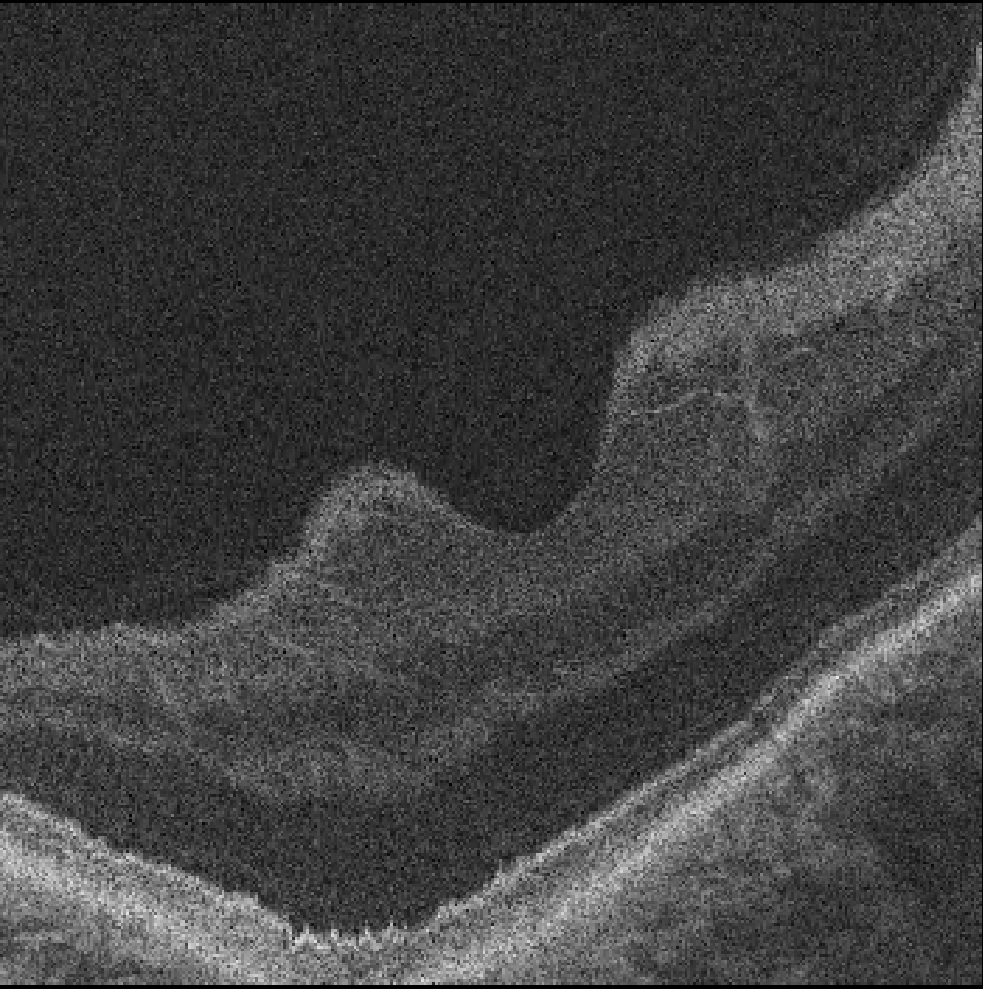}
        \caption{Low}
        \label{fig:low}
    \end{subfigure}
    \hfill
    \begin{subfigure}[b]{0.25\textwidth}
        \centering
        \includegraphics[width=\linewidth]{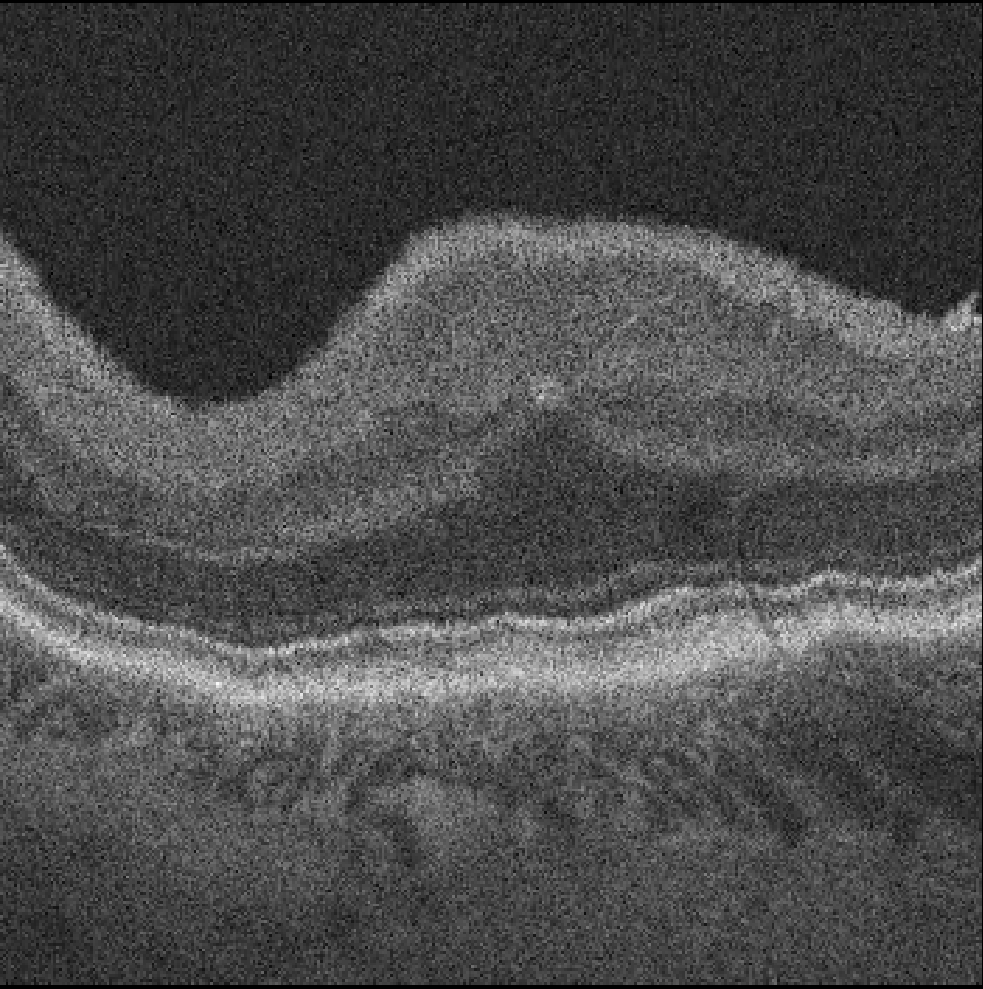}
        \caption{Borderline Low}
        \label{fig:borderline_low}
    \end{subfigure}
    \hfill
    \begin{subfigure}[b]{0.25\textwidth}
        \centering
        \includegraphics[width=\linewidth]{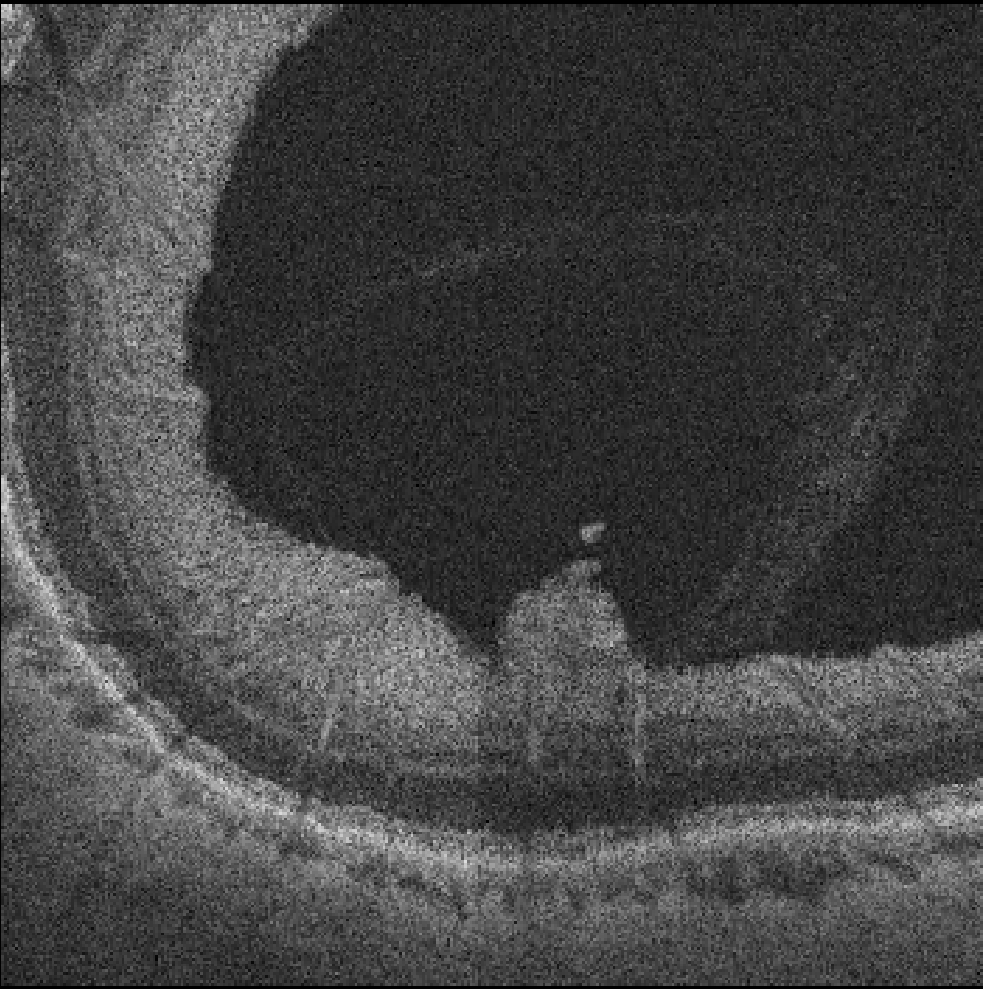}
        \caption{Normal}
        \label{fig:octa_normal}
    \end{subfigure}
    \caption{The figure illustrates OCTA instances corresponding to each level of $\text{SaO}_\text{2}$ in Prog-OCTA dataset.}
    \label{fig:octa}
\end{figure}

\vspace{-1cm}
\paragraph{\textbf{Long-Tailed Image Recognition}} Addressing the long-tailed problem in computer vision remains a challenging and critical task due to the severe class imbalance, where a few classes dominate while many others are underrepresented. This leads to biased predictions favoring the major classes, posing significant hurdles for accurate model performance. OLTR \cite{liu2019large} proposed a dynamic meta-embedding method and introduced a dataset for Open Long-Tailed Recognition (OLTR) to address imbalanced classification, few-shot learning, and open-set recognition simultaneously. Logit adjustment \cite{menon2020long} presented a method utilizing a novel balanced cross entropy (Bal-CE) loss to adjust logits based on label frequencies, encouraging a large relative margin between logits of rare and dominant labels. LiVT \cite{xu2023learning} proposed a novel method utilizing a balanced binary cross entropy (Bal-BCE) loss to address the challenges of long-tailed recognition by training Vision Transformer from scratch with long-tailed data. This approach achieves significant improvements without requiring additional data.

\begin{figure}[H]
    \centering
    \includegraphics[width=0.8\linewidth]{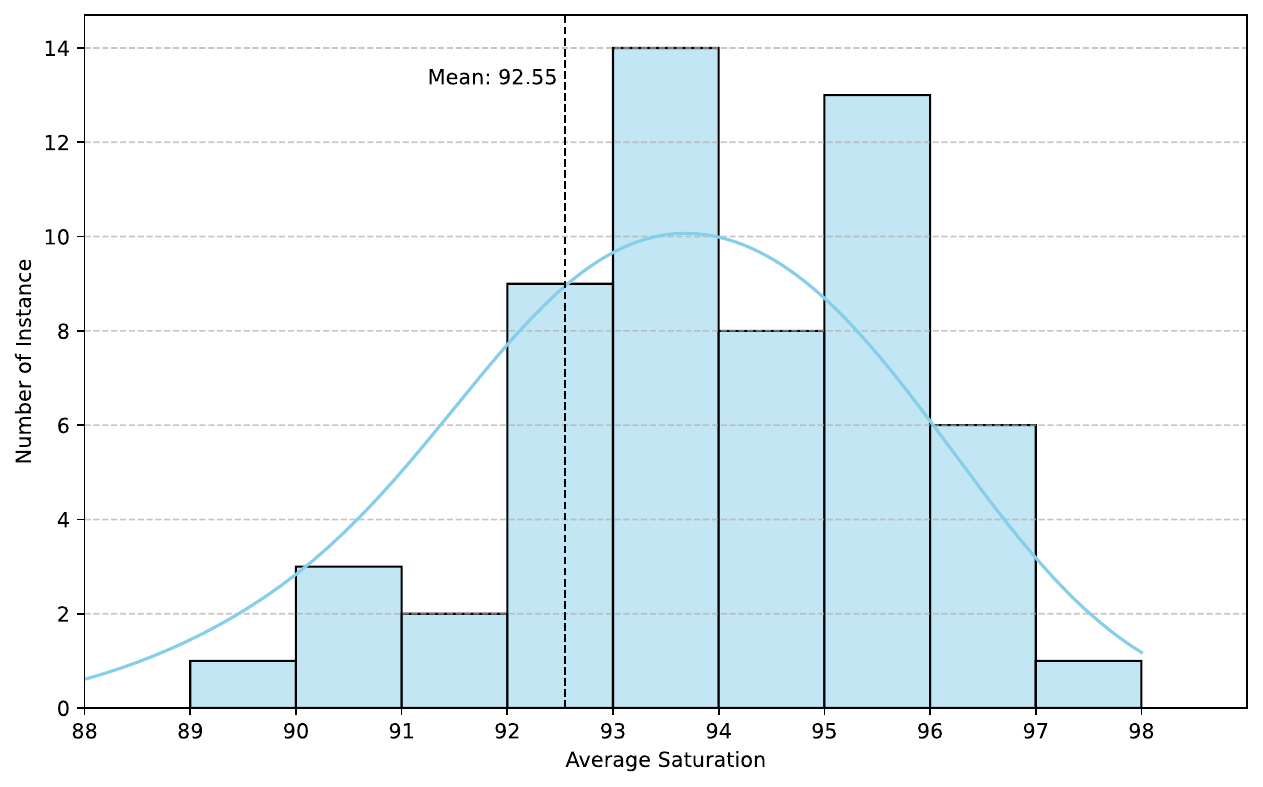} 
    \caption{The figure illustrates the $\text{SaO}_\text{2}$ value of patients in Prog-OCTA dataset, and the distribution of the dataset is imbalanced and apparently has a lower average $\text{SaO}_\text{2}$ than the normal people.}
    \label{fig:data_dist}
\end{figure}
\vspace{-1cm}

\paragraph{\textbf{OCTA in AI for Health}} Optical Coherence Tomography Angiography (OCTA) is an enhancement of OCT, enabling noninvasive imaging of ocular structures with improved visualization of retinal and anterior eye anatomy compared to conventional methods \cite{le2022optical}. OCTA-Net \cite{ma2020rose} proposed a split-based coarse-to-fine vessel segmentation method and introduced a new dataset, ROSE, focusing on retinal OCTA images, in order to address challenges in automated segmentation of retinal vessels, achieving improved performance over traditional and other deep learning methods and enabling analysis of retinal microvasculature for studying neurodegenerative diseases. Xie et al. \cite{xie2024deep} proposed a standardized OCTA analysis framework, utilizing deep learning models to extract geometrical parameters of retinal microvasculature, which enabled the comparison of these parameters among healthy controls and subjects with Alzheimer’s Disease or mild cognitive impairment (MCI), demonstrating the potential of OCTA as a diagnostic tool for these conditions. Polar-Net \cite{liu2023polar} proposed a novel deep-learning framework, involving the transformation of OCTA images from Cartesian to polar coordinates, to facilitate region-based analysis akin to the ETDRS grid commonly used in clinical practice. Additionally, it integrates clinical prior information and adapts to assess the importance of retinal regions, thereby enhancing interpretability and performance in detecting Alzheimer's disease.

\paragraph{\textbf{Oxygen Saturation Prediction}} Arterial oxygen saturation ($\text{SaO}_\text{2}$) serves as a crucial and frequently used parameter in diagnosing sleep apnea, sleep-related hypoxemia, and other types of sleep-related breathing disorders \cite{sleepfoundationWhatNormal, bark2023selanet}.  SelANet \cite{bark2023selanet} proposed a selective prediction method, using confidence scores from estimated respiratory signals to classify sleep apnea occurrence samples with high confidence and rejecting uncertain samples, achieving notable improvements in classification performance. Mathew et al. \cite{mathew2023remote} proposed a convolutional neural network based noncontact method to estimate blood oxygen saturation using smartphone cameras without direct contact with individuals, thus minimizing the risk of cross contamination. Res-SE-ConvNet \cite{mahmud2022res} proposed a residual-squeeze-excitation-attention based convolutional network method, designed to determine the severity level of hypoxemia solely using Photoplethysmography (PPG) signals, achieving high accuracy and demonstrating the potential for aiding patients in receiving faster clinical decision support. ZleepNet \cite{chaw2023zleepnet} proposed a deep convolutional neural network model for sleep apnea detection, aiming to reduce the number of sensors required for signal detection. It achieved superior accuracy compared to traditional machine learning methods on publicly available sleep datasets.

\section{Datasets}

\paragraph{\textbf{Prog-OCTA}} 

The dataset is acquired by ophthalmologists from patients who undergo ophthalmic investigations at enrollment and then every six months during monitoring. Prog-OCTA contains 57 OCTA instances aligned with fine-grained average oxygen saturation ($\text{SaO}_\text{2}$) labeled as ground truth. Since the dataset originates from patients with sleep disorders, the distribution of the dataset tends to have a lower average $\text{SaO}_\text{2}$, as illustrated in \cref{fig:data_dist}.

{\centering
\begin{minipage}{0.35\textwidth}
  \centering
  \captionof{table}{The classification of $\text{SaO}_\text{2}$ shown in this table is widely accepted by health-related research and proposed by well-established guidelines \cite{sleepfoundationWhatNormal, sateia2014international}.}
\resizebox{\textwidth}{!}{%
\begin{tabular}{c|c}
    \hline
    \textbf{$\text{SaO}_\text{2}$ (\%)} & \textbf{Classification} \\
    \hline
    96 - 100 & Normal \\
    93 - 95 & Borderline Low \\
    89 - 92 & Low \\
    \hline
\end{tabular}
}
\label{tab:sao2_classification} %
\end{minipage}
\hfill
\begin{minipage}{0.6\textwidth}
  \centering
  \includegraphics[width=0.8\linewidth]{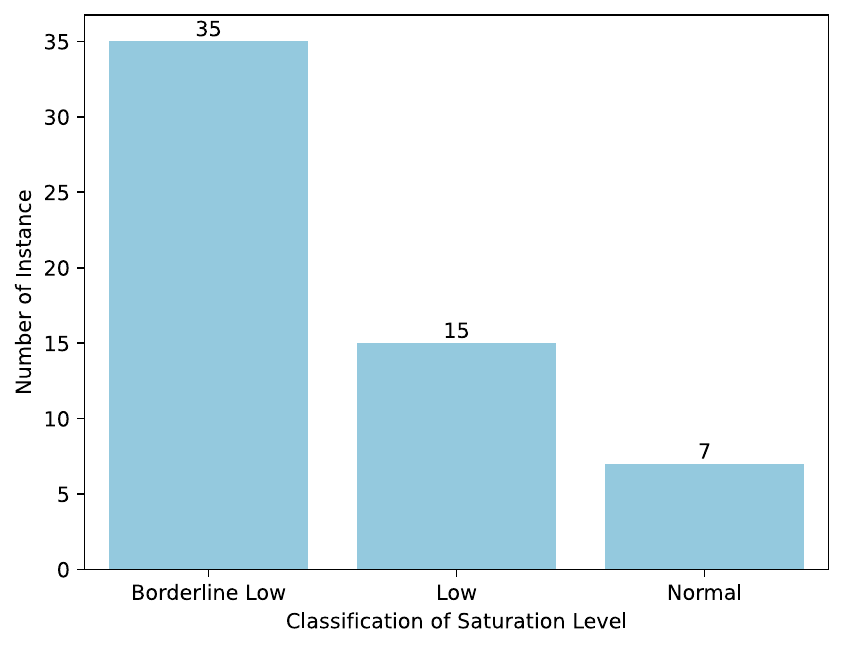} 
  \captionof{figure}{The figure shows the long-tailed and imbalanced distribution of $\text{SaO}_\text{2}$ classes in Prog-OCTA, with the borderline low class being predominant.}
  \label{fig:class_dist}
\end{minipage}
}

However, directly predicting the average $\text{SaO}_\text{2}$ is not straightforward, and may not adequately reflect patients' saturation status for sleep disorder diagnosis. Additionally, it may lack representation power and robustness based solely on the instance number among each discrete $\text{SaO}_\text{2}$ value. Hence, we convert the discrete $\text{SaO}_\text{2}$ values to the oxygen saturation levels according to widely accepted referential standards in \cref{tab:sao2_classification} \cite{sleepfoundationWhatNormal, sateia2014international}.

After refactoring the $\text{SaO}_\text{2}$ values into classes according to \cref{tab:sao2_classification}, we have organized the dataset into three classes. The distribution of each class is depicted in \cref{fig:class_dist}. This illustrates that the distribution of $\text{SaO}_\text{2}$ is severely long-tailed and imbalanced. The borderline low categories have significantly more instances than the combination of the other two classes, reflecting the actual $\text{SaO}_\text{2}$ status observed in the dataset, and an example for each class is provided in \cref{fig:octa}.

\begin{figure}[H]
    \centering
    \begin{subfigure}[b]{0.24\textwidth}
        \centering
        \includegraphics[width=\linewidth]{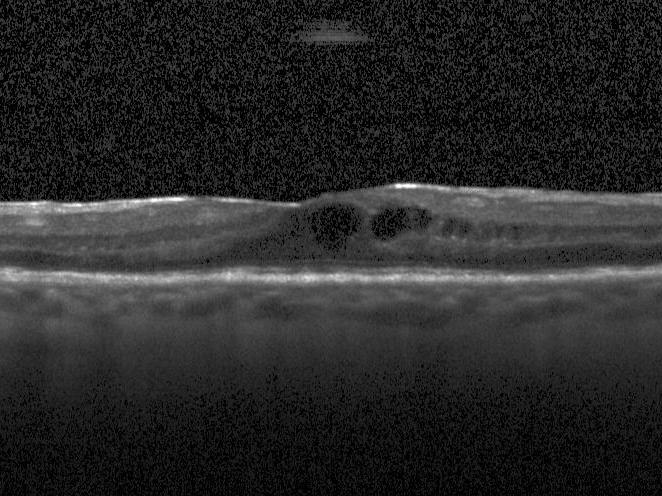}
        \caption{DME}
        \label{fig:dme}
    \end{subfigure}
    \hfill
    \begin{subfigure}[b]{0.24\textwidth}
        \centering
        \includegraphics[width=\linewidth]{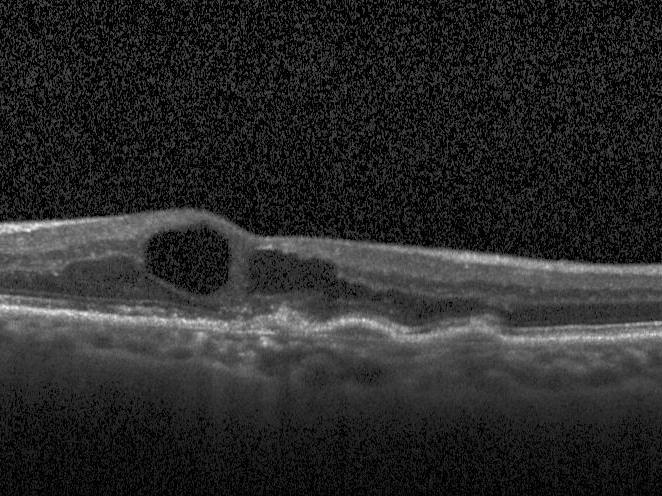}
        \caption{CNV}
        \label{fig:cnv}
    \end{subfigure}
    \hfill
    \begin{subfigure}[b]{0.24\textwidth}
        \centering
        \includegraphics[width=\linewidth]{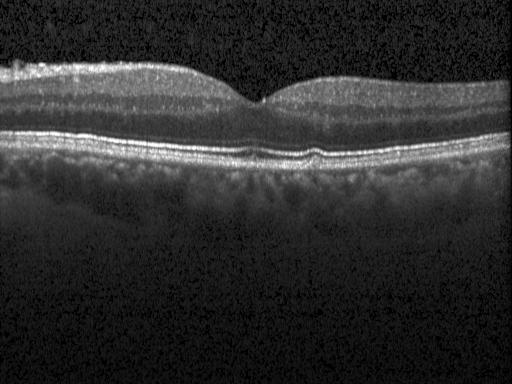}
        \caption{Drusen}
        \label{fig:drusen}
    \end{subfigure}
    \hfill
    \begin{subfigure}[b]{0.24\textwidth}
        \centering
        \includegraphics[width=\linewidth]{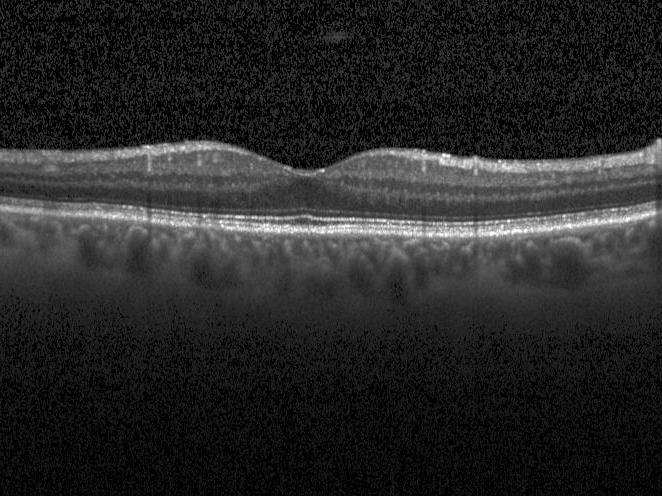}
        \caption{Normal}
        \label{fig:normal}
    \end{subfigure}
    \caption{The figures depict four categories of OCT included in Kermany v3 \cite{kermany2018identifying}: Diabetic macular edema (DME), characterized by fluid accumulation in the macula due to diabetes; CNV (Choroidal Neovascularization), involving abnormal growth of blood vessels in the retina; Drusen, indicated by the accumulation of deposits comprised of lipids and proteins in the retina; and normal instances.}
    \label{fig:kermany}
\end{figure}

\paragraph{\textbf{Kermany v3}} The Kermany database v3 \cite{kermany2018identifying} is a publicly available dataset comprising 109,312 structural retinal OCT images. It is widely used in OCT-related clinical studies and AI for health research \cite{talcott2024automated, george2024two, song2024roboctnet}. It comprises four different categories: diabetic macular edema (DME), choroidal neovascularization (CNV), Drusen, and normal instances. The dataset distribution is shown in \cref{table:kermany_dist} and \cref{fig:kermany_dist}, and an example for each class is provided in \cref{fig:kermany}. The Kermany v3 dataset was subsequently utilized for post-training our JointViT model, aimed at enhancing the inherent representation derived from COVID-ViT \cite{gao2022covid} pretraining.

{\centering
\begin{minipage}{0.45\textwidth}
  \centering
  \captionof{table}{The table displays the number of instances in each class within the training set and testing set of the Kermany database v3 \cite{kermany2018identifying}.}
\resizebox{\textwidth}{!}{%
\begin{tabular}{cccccc}
\midrule
\multicolumn{1}{c}{\textbf{SETS}} & \multicolumn{5}{|c}{\textbf{CLASSES}}   \\
\midrule
  & CNV & DME & Drusen & Normal & \textbf{Total} \\ \midrule 
Train & 37206 & 11349 & 8617 & 51140 & \textbf{108312} \\
Test & 250 & 250 & 250 & 250 & \textbf{1000} \\
\midrule
\end{tabular}%
}
\label{table:kermany_dist} %
\end{minipage}
\hfill
\begin{minipage}{0.45\textwidth}
  \centering
  \includegraphics[width=0.8\linewidth]{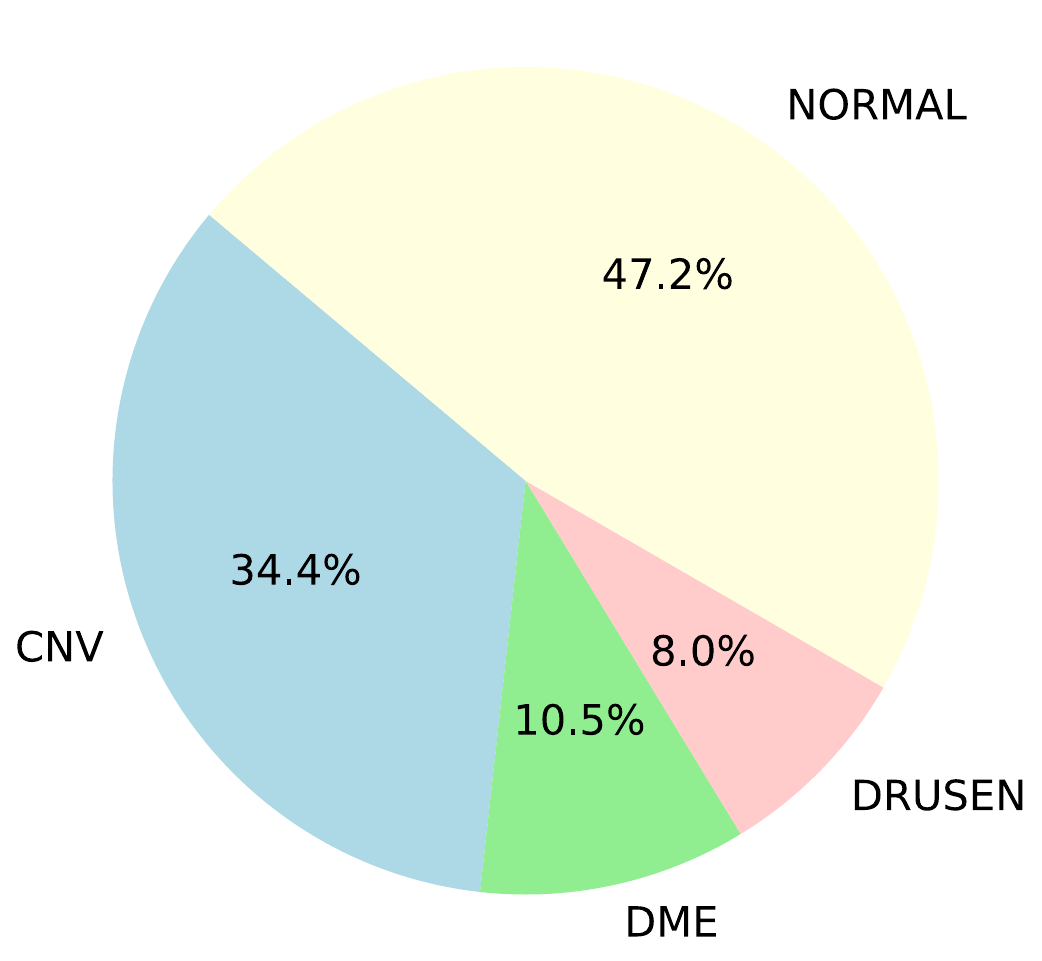} 
  \captionof{figure}{The pie chart shows the proportion of different classes of OCT in the Kermany database v3 \cite{kermany2018identifying}.}
  \label{fig:kermany_dist}
\end{minipage}
}

\section{Methodology}

\paragraph{\textbf{Balancing Augmentation}}

To address the challenges posed by the long-tailed distribution and severe class imbalance in the Prog-OCTA dataset and Kermany v3 dataset \cite{kermany2018identifying}, we propose the integration of \textit{\textbf{balancing augmentation}} into the data preprocessing pipeline. Unlike conventional data balancing techniques such as upsampling and downsampling \cite{zhang2024deep}, which often introduce biases or compromise critical data information while achieving class balance, balancing augmentation employs \textit{\textbf{random cropping}}, \textit{\textbf{flipping}}, and \textit{\textbf{rotation}} at arbitrary angles. This approach introduces increased variability into the original dataset during the balancing process without introducing additional biases.

The balancing augmentation technique is implemented on the minority classes of both datasets, augmenting them to match the number of instances in the majority class, as illustrated in \cref{fig:jointvit}, as part of the data preprocessing phase. This augmentation process serves to rectify imbalances within the dataset, particularly addressing long-tailed distribution issues from an input perspective. By ensuring a balanced representation of classes, the model's training process becomes more robust and less biased towards the majority class, thus enhancing its ability to generalize effectively across diverse data distributions. Employing this technique ultimately fosters improved model performance and contributes to more equitable and reliable predictions across all classes.

\paragraph{\textbf{JointViT Backbone}}

We utilize a Vision Transformer (ViT) \cite{dosovitskiy2020image} as our backbone, combined with a linear classification head. The ViT encoder is pretrained on COVID-ViT \cite{gao2022covid} and further fine-tuned via transfer learning for our downstream task. To augment the learned representation of ViT, we additionally perform post-training on the Kermany v3 OCT dataset \cite{kermany2018identifying} as an initialization for our $\text{SaO}_\text{2}$ prediction task on Prog-OCTA.

During post-training, the augmented OCT images in the Kermany v3 dataset \cite{kermany2018identifying} are divided into $k$ patches of size $16 \times 16$ and processed through self-attention layers, incorporating patch and position embeddings. Unlike OCT images, OCTA images in Prog-OCTA are 3D volumes. Therefore, they are initially converted into 2D slices, then they are performed balancing augmentation and divided into patches, which is similar to the post-training process. The reason behind converting 3D OCTA volumes into 2D slices is twofold. Firstly, there is a lack of pretrained models designed specifically for 3D Vision Transformers \cite{zhu20233d} compared to well-established pretraining models for 2D Vision Transformers \cite{he2022masked, oquab2023dinov2}. Secondly, utilizing raw 3D volumes as input would significantly increase computational complexity and demand greater computational resources compared to using 2D slices.

Unlike the vanilla Vision Transformers, which uses image categories as supervision, we employ both $\text{SaO}_\text{2}$ categories $Y_{\text{cls}}$ and exact $\text{SaO}_\text{2}$ values $Y_{\text{val}}$ as joint supervision. For classification, we utilize binary cross-entropy loss $L_{\text{bce}}$, which outperforms ordinary cross-entropy loss when combined with Vision Transformers \cite{touvron2022deit, xu2023learning}. For regression, we adopt mean squared error loss $L_{\text{mse}}$ for supervision. The weight of each loss in the joint loss is controlled by coefficient $\lambda$, as shown in \cref{eq:jointloss}. Assuming the model is denoted as $T$ and the input image as $X$, the joint loss $L$ can be expressed as follows.

\begin{equation}
    L = \lambda L_{\text{bce}}(T(X), Y_{\text{cls}}) + (1 - \lambda)L_{\text{mse}}(T(X), Y_{\text{val}})
    \label{eq:jointloss}
\end{equation}

Additionally, while our MLP head predicts both classes and values for $\text{SaO}_\text{2}$, our approach differs from a multitasking model that handles both classification and regression tasks. We choose to focus solely on predicting the $\text{SaO}_\text{2}$ classes as the desired output. It is worth noting that accurately predicting precise $\text{SaO}_\text{2}$ values with limited data is impractical. Instead, we assign minimal weight to the mean squared error (MSE) loss in the joint loss function to prioritize the performance of classification.

\section{Experiments}

\subsection{Evaluation Matrices}

For a fair comparison, we used the test set accuracy to evaluate the overall classification performance of the methods, meanwhile using sensitivity and specificity to assess how the model handles both minority and majority classes in the long-tailed Prog-OCTA dataset.

\begin{table}[H]
\centering
\resizebox{0.85\textwidth}{!}{
\begin{tabular}{c|ccc}
\toprule
Models & Test Acc. $\uparrow$ & Sensitivity $\uparrow$ & Specificity $\uparrow$ \\ \midrule 
M3T \cite{jang2022m3t} & $61.40^{\pm2.48}$ & $33.33^{\pm0.00}$ & $66.67^{\pm0.00}$ \\
I3D \cite{carreira2017quo} & $61.40^{\pm2.48}$ & $33.33^{\pm0.00}$ & $66.67^{\pm0.00}$ \\
MedViT \cite{manzari2023medvit}& $61.40^{\pm2.48}$ & $33.33^{\pm0.00}$ & $66.67^{\pm0.00}$ \\
FCNlinksCNN \cite{qiu2020development} & $57.89^{\pm7.44}$ & $31.31^{\pm2.86}$ & $64.96^{\pm2.42}$ \\
MedicalNet (3D ResNet-10) \cite{chen2019med3d} & $63.16^{\pm0.00}$ & $37.96^{\pm3.47}$ & $68.95^{\pm1.72}$ \\
MedicalNet (3D ResNet-18) \cite{chen2019med3d} & $63.16^{\pm4.30}$ & $40.74^{\pm8.59}$ & $71.62^{\pm5.52}$ \\
MedicalNet (3D ResNet-34) \cite{chen2019med3d} & $61.40^{\pm2.48}$ & $33.33^{\pm0.00}$ & $66.67^{\pm0.00}$ \\
MedicalNet (3D ResNet-50) \cite{chen2019med3d} & $61.40^{\pm2.48}$ & $33.33 ^{\pm0.00}$ & $66.67^{\pm0.00}$ \\
MedicalNet (3D ResNet-101) \cite{chen2019med3d} & $61.40^{\pm2.48}$ & $33.33 ^{\pm0.00}$ & $66.67^{\pm0.00}$ \\
MedicalNet (3D ResNet-152) \cite{chen2019med3d} & $61.40^{\pm2.48}$ & $33.33 ^{\pm0.00}$ & $66.67^{\pm0.00}$ \\
MedicalNet (3D ResNet-200) \cite{chen2019med3d} & $61.40^{\pm2.48}$ & $33.33 ^{\pm0.00}$ & $66.67^{\pm0.00}$ \\
3D DenseNet-121 \cite{huang2017densely} & $66.67^{\pm2.48}$ & $45.37^{\pm3.46}$ & $72.58 ^{\pm0.82}$ \\
3D DenseNet-161 \cite{huang2017densely} & $66.67^{\pm2.48}$ & $47.22^{\pm2.27}$ & $75.06^{\pm2.14}$ \\
3D DenseNet-169 \cite{huang2017densely} & $66.67^{\pm2.48}$ & $43.70^{\pm7.97}$ & $72.90^{\pm2.37}$ \\
3D DenseNet-201 \cite{huang2017densely} & $64.91^{\pm4.96}$ & $39.81^{\pm4.72}$ & $71.47^{\pm3.40}$ \\
COVID-ViT \cite{gao2022covid} & $61.40^{\pm6.40}$ & $36.63^{\pm7.76}$ & $61.66^{\pm8.13}$ \\
\textbf{JointViT (Ours)} & $\textbf{70.17}^{\pm8.90}$ & $\textbf{62.51}^{\pm8.44}$ & $\textbf{82.83}^{\pm7.07}$ \\
\midrule
\end{tabular}
}
\caption{We compared our proposed \textbf{JointViT} on Prog-OCTA dataset with well-established 2D and 3D methods which are widely used in medical imaging recognition. The results demonstrate that our method significantly outperforms others in predicting saturation levels using imbalanced OCTA data.}
\label{table:1}
\end{table}

\subsection{Comparative Studies}

To thoroughly evaluate our proposed JointViT, we conducted comprehensive experiments comparing it with other well-established methods, as demonstrated in \cref{table:1}. Inspired by the principle of K-fold cross-validation, we divided the Prog-OCTA dataset randomly into 3 folds. Each time, we used two folds for training and one fold for testing, repeating the experiments three times for each model. This approach ensures a fair and robust evaluation, particularly when dealing with imbalanced and limited data. The experiments are conducted with an optimal $\lambda$ set to $0.99$, and the ablation of different values of $\lambda$ has been conducted in the following section. The results displayed in \cref{table:1} demonstrate that our performance significantly outperforms other methods by up to 12.28\% in overall accuracy. This achievement represents the state-of-the-art performance in modeling $\text{SaO}_\text{2}$ using long-tailed OCTA.

Given the long-tailed nature of the data, all models exhibit a pattern of higher specificity than sensitivity. This is because they excel at identifying common classes but face challenges in recognizing less frequent ones. However, our approach demonstrates not only higher specificity compared to other methods but, more crucially, significantly higher sensitivity. This indicates that our model not only effectively handles major classes but also possesses superior capability in addressing long-tailed classes compared to compared methods.

\subsection{Ablation Studies}

\begin{figure}[H]
    \centering
    \includegraphics[width=0.7\linewidth]{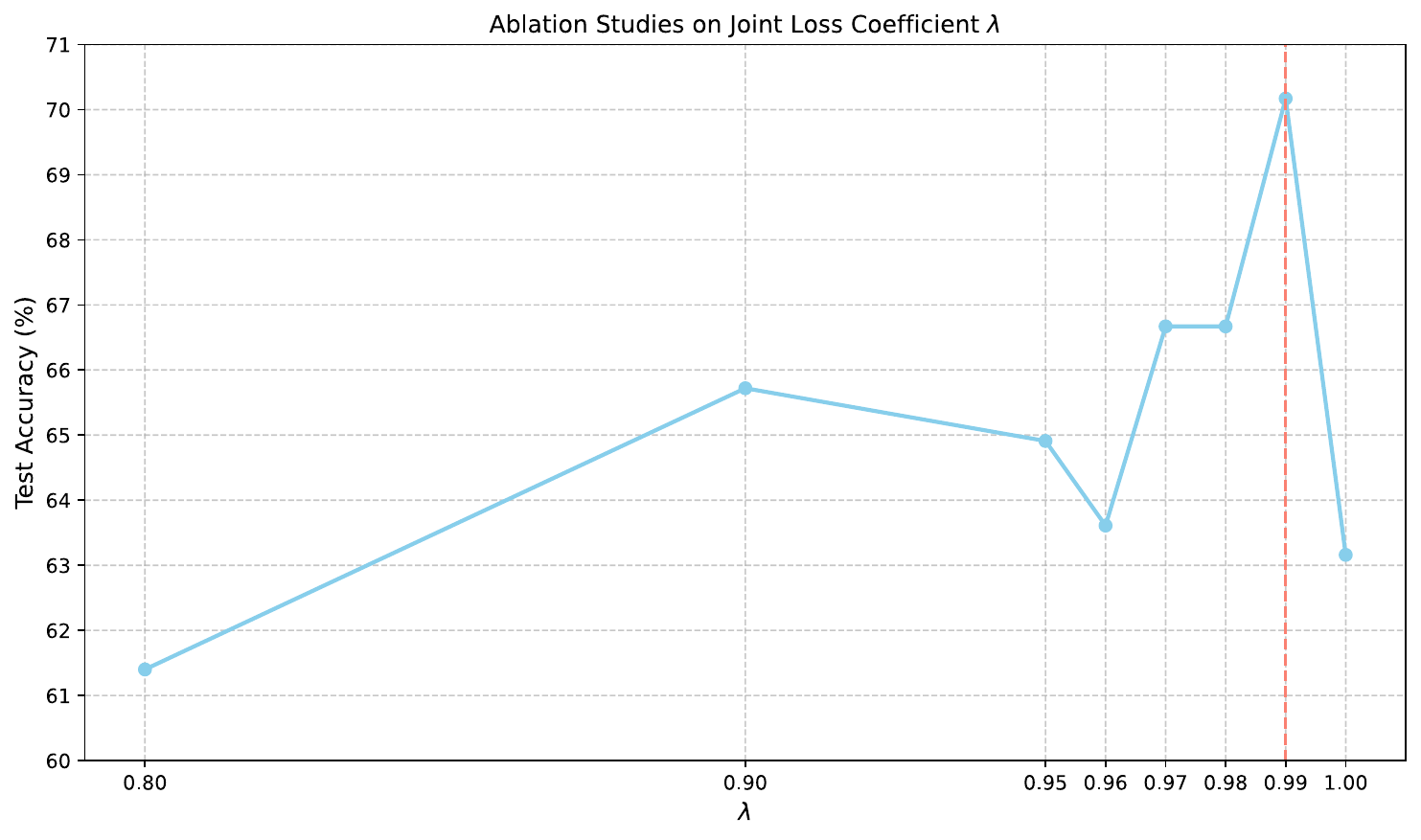} 
    \caption{The figure visualizes the relationship between joint loss coefficient ($\lambda$) values and overall model performance, which indicates the optimal $\lambda$ value is 0.99.}
    \label{fig:lambda_ablation}
\end{figure}

We performed experiments by varying the joint loss coefficient $\lambda$ to examine its influence on both the joint loss and the overall performance of the model. The outcomes are detailed in  \cref{table:lambda_ablation} and illustrated in \cref{fig:lambda_ablation}. Notably, the results indicate that the model achieves optimal performance when $\lambda$ is set to 0.99.

\begin{table}[H]
\centering
\resizebox{0.8\textwidth}{!}{
\begin{tabular}{c|ccc}
\toprule
Models & Test Acc. $\uparrow$ & Sensitivity $\uparrow$ & Specificity $\uparrow$ \\ \midrule 
COVID-ViT \cite{gao2022covid} & $61.40^{\pm6.40}$ & $36.63^{\pm7.76}$ & $61.66^{\pm8.13}$ \\
JointViT ($\lambda$ = 0.8) & $61.40^{\pm4.71}$ & $33.35^{\pm5.44}$ & $67.41^{\pm7.27}$ \\
JointViT ($\lambda$ = 0.9) & $65.72^{\pm2.72}$ & $48.89^{\pm4.20}$ & $66.27^{\pm3.13}$ \\
JointViT ($\lambda$ = 0.95) & $64.91^{\pm4.21}$ & $54.67^{\pm6.11}$ & $70.31^{\pm3.82}$ \\
JointViT ($\lambda$ = 0.96) & $63.61^{\pm6.67}$ & $50.43^{\pm4.51}$ & $76.20^{\pm2.10}$ \\
JointViT ($\lambda$ = 0.97) & $66.67^{\pm2.46}$ & $57.22^{\pm7.31}$ & $64.82^{\pm6.20}$ \\
JointViT ($\lambda$ = 0.98) & $66.67^{\pm4.50}$ & $51.32^{\pm3.42}$ & $70.42^{\pm6.26}$ \\
JointViT ($\lambda$ = 1) & $63.16^{\pm0.00}$ & $35.92^{\pm3.21}$ & $76.81^{\pm2.29}$ \\
\textbf{JointViT ($\lambda$ = 0.99)} & $\textbf{70.17}^{\pm8.90}$ & $\textbf{62.51}^{\pm8.44}$ & $\textbf{82.83}^{\pm7.07}$ \\
\midrule
\end{tabular}
}
\caption{Ablation results in the table shows the impact of varying joint loss coefficient ($\lambda$) values on model performance, which indicates when $\lambda$ is set to be 0.99, the model achieved optimal performance.}
\label{table:lambda_ablation}
\end{table}

We have additionally integrated ablation analysis concerning the \textit{joint loss} within the proposed JointViT framework. This involved substituting our joint loss with a modified binary cross entropy loss, balanced BCE loss (\textit{Bal-BCE}) \cite{xu2023learning}, which tailors specifically for addressing the challenges inherent in long-tailed image recognition problems. The outcomes of this ablation study are presented in \cref{table:loss_ablation}, which shows that our joint loss significantly outperforms well-established Bal-BCE loss designs tailored for long-tailed image recognition scenarios, elucidating that our proposed joint loss formulation stands as the optimal selection for tackling the long-tailed distribution of Prog-OCTA. This underscores the efficacy and superiority of our devised joint loss scheme in addressing the intricacies of long-tailed image recognition tasks, particularly within the OCTA recognition domain. To clarify, in the ablation study of JointViT without the joint loss, only the BCE loss is utilized. In the case of the ablation study on JointViT with Bal-BCE loss, when $\lambda = 1$, the loss exclusively consists of the Bal-BCE loss, and setting $\lambda = 0.99$ substitutes the BCE loss with the Bal-BCE loss in the joint loss function.

\vspace{-0.5cm}
\begin{table}[H]
\centering
\resizebox{0.9\textwidth}{!}{
\begin{tabular}{c|ccc}
\toprule
Models & Test Acc. $\uparrow$ & Sensitivity $\uparrow$ & Specificity $\uparrow$ \\ \midrule 
COVID-ViT \cite{gao2022covid} & $61.40^{\pm6.40}$ & $36.63^{\pm7.76}$ & $61.66^{\pm8.13}$ \\
JointViT w/o joint loss & $63.16 ^{\pm0.00}$ & $35.92^{\pm3.21}$ & $76.81^{\pm2.29}$ \\
JointViT w/ Bal-BCE ($\lambda$ = 1) & $61.40^{\pm2.48}$ & $20.47^{\pm0.82}$ & $53.80^{\pm0.83}$ \\
JointViT w/ Bal-BCE ($\lambda$ = 0.99) & $64.91^{\pm2.48}$ & $45.63^{\pm17.55}$ & $71.08^{\pm11.96}$ \\
\textbf{JointViT w/ joint loss ($\lambda$ = 0.99)} & $\textbf{70.17}^{\pm8.90}$ & $\textbf{62.51}^{\pm8.44}$ & $\textbf{82.83}^{\pm7.07}$ \\
\midrule
\end{tabular}
}
\caption{The table presents the results of conducting ablations involving the integration of different loss functions within the joint loss framework. The results indicate that our jointly designed loss function, which combines BCE and MSE losses, consistently achieves superior performance compared to alternative configurations.}
\label{table:loss_ablation}
\end{table}
\vspace{-0.7cm}

In our comparative studies, we further explored the efficacy of utilizing the other backbones through ablation analysis, especially compared with the second best backbone in comparative studies. The findings, as depicted in \cref{table:backbone}, underscore that our approach leveraging a ViT backbone continues to exhibit superior performance compared to alternative backbone architectures. This reaffirms the efficacy of our method in achieving optimal results across diverse backbone configurations.
\vspace{-0.5cm}

\begin{table}[H]
\centering
\resizebox{\textwidth}{!}{
\begin{tabular}{c|ccc}
\toprule
Models & Test Acc. $\uparrow$ & Sensitivity $\uparrow$ & Specificity $\uparrow$ \\ \midrule 
2D DenseNet-161 w/ bal-aug \& joint loss & $59.65^{\pm4.96}$ & $26.26^{\pm7.36}$ & $60.17^{\pm8.17}$ \\
3D DenseNet-161 w/ bal-aug \& joint loss & $55.09^{\pm3.46}$ & $44.82^{\pm6.89}$ & $68.37^{\pm10.46}$ \\
\textbf{JointViT} & $\textbf{70.17}^{\pm8.90}$ & $\textbf{62.51}^{\pm8.44}$ & $\textbf{82.83}^{\pm7.07}$ \\
\midrule
\end{tabular}
}
\caption{We further explored the efficacy of employing the second-best backbone in comparative studies. Our findings reveal that our approach which used a ViT backbone maintains its position as the top-performing method.}
\label{table:backbone}
\end{table}
\vspace{-0.5cm}

\vspace{-0.7cm}

We have additionally assessed the effects of post-training on our training approach. The outcomes illustrated in \cref{table:post} indicate that incorporating post-training with OCT images from the Kermany database v3 \cite{kermany2018identifying} yields notably superior performance compared to the absence of post-training. This observation underscores the efficacy of employing post-training as a valuable initialization step for subsequent OCTA recognition tasks.

\vspace{-0.5cm}
\begin{table}[H]
\centering
\resizebox{0.85\textwidth}{!}{
\begin{tabular}{c|ccc}
\toprule
Models & Test Acc. $\uparrow$ & Sensitivity $\uparrow$ & Specificity $\uparrow$ \\ \midrule 
COVID-ViT \cite{gao2022covid} & $61.40^{\pm6.40}$ & $36.63^{\pm7.76}$ & $61.66^{\pm8.13}$ \\
JointViT w/o post-training & $57.89^{\pm0.00}$ & $37.45^{\pm7.76}$ & $65.85^{\pm4.40}$ \\
\textbf{JointViT w/ post-training} & $\textbf{70.17}^{\pm8.90}$ & $\textbf{62.51}^{\pm8.44}$ & $\textbf{82.83}^{\pm7.07}$ \\
\midrule
\end{tabular}
}
\caption{The table demonstrates the significant performance improvement achieved by incorporating post-training with OCT images from the Kermany database v3 \cite{kermany2018identifying}, compared to the absence of post-training, thereby validating its efficacy as an initialization step for downstream OCTA recognition tasks.}
\label{table:post}
\end{table}
\vspace{-0.8cm}

Eventually, we proceeded to incorporate ablations of our data preprocessing technique, specifically focusing on balancing augmentation (bal-aug). The results, as shown in \cref{table:bal-aug_ablation}, underscore the efficacy of balancing augmentation. Notably, our model demonstrates remarkable performance improvement with the inclusion of balancing augmentation.

\vspace{-0.7cm}
\begin{table}[H]
\centering
\resizebox{0.8\textwidth}{!}{
\begin{tabular}{c|ccc}
\toprule
Models & Test Acc. $\uparrow$ & Sensitivity $\uparrow$ & Specificity $\uparrow$ \\ \midrule 
COVID-ViT \cite{gao2022covid} & $61.40^{\pm6.40}$ & $36.63^{\pm7.76}$ & $61.66^{\pm8.13}$ \\
JointViT w/o bal-aug & $64.91^{\pm2.48}$ & $45.63^{\pm17.55}$ & $71.08^{\pm11.96}$ \\
\textbf{JointViT w/ bal-aug} & $\textbf{70.17}^{\pm8.90}$ & $\textbf{62.51}^{\pm8.44}$ & $\textbf{82.83}^{\pm7.07}$ \\
\midrule
\end{tabular}
}
\caption{The table showcases the efficacy of balancing augmentation in enhancing model performance, with notable improvements observed in various metrics.}
\label{table:bal-aug_ablation}
\end{table}

\section{Discussion and Conclusion}

Oxygen saturation level ($\text{SaO}_\text{2}$) significantly impacts health, particularly indicating conditions like sleep-related hypoxemia, hypoventilation, sleep apnea, and other related disorders. However, continuous monitoring of $\text{SaO}_\text{2}$ is time-consuming and subject to variation due to patients' status instability. On the other hand, Optical coherence tomography angiography (OCTA) images excel in speed and effectively screening eye-related lesions, showing promise in assisting with the diagnosis of sleep-related disorders. Therefore, we propose a novel method called JointViT, which incorporates a joint loss utilizing both $\text{SaO}_\text{2}$ values and categories for supervision. Additionally, we employ a balancing augmentation technique during data preprocessing to handle the long-tail nature of OCTA datasets. Our method significantly outperformed other established medical imaging recognition methods, paving the way for the future utilization of OCTA in diagnosing sleep-related disorders.


\end{document}